\documentclass[letterpaper, 10 pt, conference]{ieeeconf}  % Comment this line out if you need a4paper

\IEEEoverridecommandlockouts                              % This command is only needed if 
\usepackage{cite}
\usepackage{amsmath,amssymb,amsfonts}
\usepackage{algorithmic}
\usepackage{graphicx}
\usepackage{textcomp}
\usepackage{xcolor}
\def\BibTeX{{\rm B\kern-.05em{\sc i\kern-.025em b}\kern-.08em
		T\kern-.1667em\lower.7ex\hbox{E}\kern-.125emX}}
\usepackage{tabularx}
\usepackage{pifont}
\usepackage{booktabs}
\usepackage{multirow}
\usepackage{comment}

\title{FIMP: Future Interaction Modeling for Multi-Agent Motion Prediction
}

\author{Sungmin Woo, Minjung Kim, Donghyeong Kim, Sungjun Jang, Sangyoun Lee$^{*}$\\
\thanks{*Corresponding author.}% <-this % stops a space
\thanks{The authors are with the School of Electrical and Electronic Engineering, Yonsei University, Seoul, Korea.
}}

\begin{document}

\maketitle
\thispagestyle{empty}
\pagestyle{empty}

%%%%%%%%%%%%%%%%%%%%%%%%%%%%%%%%%%%%%%%%%%%%%%%%%%%%%%%%%%%%%%%%%%%%%%%%%%%%%%%%
\begin{abstract}
Multi-agent motion prediction is a crucial concern in autonomous driving, yet it remains a challenge owing to the ambiguous intentions of dynamic agents and their intricate interactions. Existing studies have attempted to capture interactions between road entities by using the definite data in history timesteps, as future information is not available and involves high uncertainty. However, without sufficient guidance for capturing future states of interacting agents, they frequently produce unrealistic trajectory overlaps. In this work, we propose Future Interaction modeling for Motion Prediction (FIMP), which captures potential future interactions in an end-to-end manner. FIMP adopts a future decoder that implicitly extracts the potential future information in an intermediate feature-level, and identifies the interacting entity pairs through future affinity learning and top-$k$ filtering strategy. Experiments show that our future interaction modeling improves the performance remarkably, leading to superior performance on the Argoverse motion forecasting benchmark.
\end{abstract}

%%%%%%%%%%%%%%%%%%%%%%%%%%%%%%%%%%%%%%%%%%%%%%%%%%%%%%%%%%%%%%%%%%%%%%%%%%%%%%%%
\section{Introduction}
Accurate motion prediction is one of the essential requirements for safe and robust autonomous driving. Anticipating the near future enables a thorough understanding of the surrounding contexts and serves as the fundamental grounds for automated decision-making. However, it remains a challenge because the ambiguous intentions of dynamic agents involve high uncertainty and their behaviors are considerably affected by environmental constraints, such as kinematic states of neighboring agents, map topology and traffic rules. It is not straightforward to precisely capture all potential interactions between those factors in complex multi-agent scenarios.

Recent deep learning approaches have proposed diverse solutions for interaction modeling. Raster-based methods~\cite{chai2019multipath, cui2019multimodal, gilles2021home} rasterize the scene information as a multi-channel image from a top-down view, and encode the local interactions by using off-the-shelf 2D convolutional neural networks (CNNs). Graph-based methods~\cite{zhou2022hivt, gao2020vectornet, liang2020learning} employ a vectorized representation that organizes data as polylines, and apply graph neural networks (GNNs) to learn the flow of information between nodes. Attention mechanisms are also extensively utilized in numerous methods~\cite{zhou2022hivt, gao2020vectornet, liang2020learning, ye2021tpcn, wang2022ltp, liu2021multimodal, ngiam2021scene, salzmann2020trajectron++} to better capture long-term interactions by modeling the relationships between entities (e.g., trajectory waypoints, lane segments) in spatial and temporal aspects. 

\begin{figure}[t]
	\begin{center}
		\includegraphics[width=\linewidth]{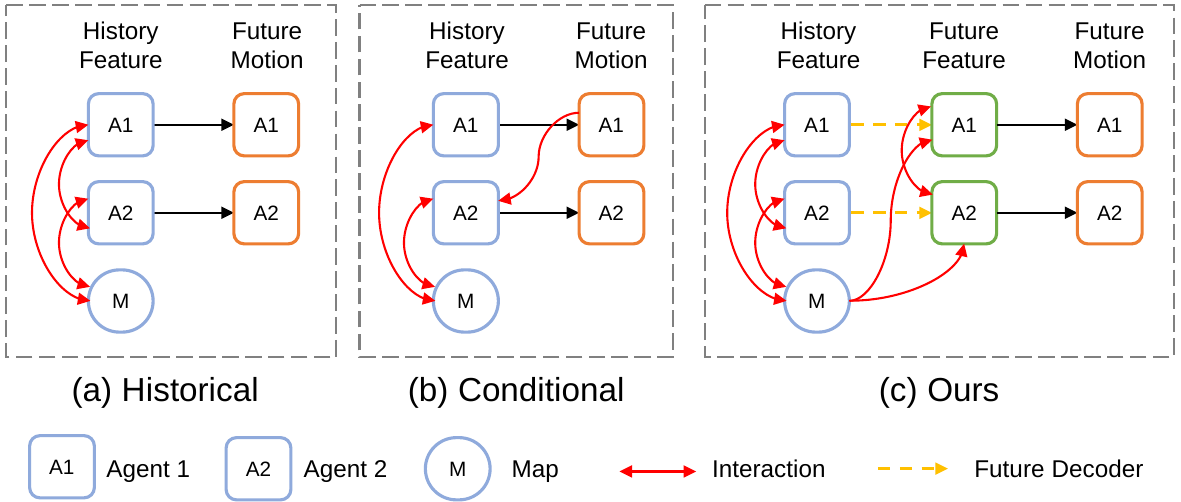}
	\end{center}
	\vspace{-0.5cm}
	\caption{Overview of interaction modeling in motion prediction. (a) Observed-historical-information-based interaction. (b) Estimated high-level future-states-based conditional prediction. (c) Our feature-level potential future information based interaction.
	}
	\vspace{-0.5cm}
	\label{fig:1}
\end{figure}
%VectorNet~\cite{} . LaneGCN, latest methods~\cite{}, spatial.temporal.entity -level (LTP - interaction modeling). 

As the available data given for motion prediction is from past timesteps, the interaction modeling of most approaches is naturally designed to focus on the observed historical information. The interaction is usually computed between entity features extracted from the observed data (i.e., history feature) as illustrated in Figure~\ref{fig:1} (a). However, this architecture lacks sufficient guidance for modeling interactions in future timesteps as message passing between predicted future motions is not significantly focused and performed. Some works~\cite{salzmann2020trajectron++, sun2022m2i, tolstaya2021identifying, khandelwal2020if} address this issue by solving the conditional prediction, which is to predict the motion conditioned on the estimated future states of interacting agents (Figure~\ref{fig:1} (b)). This intuitive approach can take into account future interactions directly, but there remain three drawbacks. (1) As explicit high-level future information (e.g., future trajectory, goal point) is required, the interaction modeling heavily relies on the pre-estimated future motions of other agents. The accuracy of pre-estimation determines the reliability of considered interaction. (2) The conditional prediction often neglects the mutual influence between interacting agents. It lacks the ability to predict the motions of multi-agents jointly when both behaviors of agents are influenced by each other. (3) Owing to the additional procedure to estimate and refine the motion, the entire prediction process is inefficient and not suitable for real-time applications.

To alleviate these problems, we propose Future Interaction modeling for Motion Prediction (FIMP), which learns to capture a future interaction in an end-to-end manner. Instead of using the pre-estimated high-level future information, FIMP utilizes the features that implicitly contain the potential future information (i.e., future feature). As shown in Figure~\ref{fig:1} (c), we decouple future features from the history feature for each agent, enabling to model the mutual future interaction without high-level cues as well as history interaction.

%In our proposed approach, we address two issues. First, future feature should be essentially extracted aside from history feature and involve potential future information. 

Specifically, we derive future features by adopting an intermediate future decoder comprised of a multi-head projection layer and a gated recurrent unit (GRU)~\cite{cho2014properties}. The multi-head projection layer extracts disparate future mode embeddings from the history feature and GRU temporalizes each mode embedding into specific future time zones, which are the small time chunks split from entire prediction time. The acquired zone-wise future feature is then optimized to precisely determine when and where agents will be in the corresponding time period by learning. In addition, pairs of interacting agents are identified without prior knowledge of the explicit future states of agents in our approach. Instead, FIMP learns to extract affinity between future features of agents and selects the agent pairs with top-$k$ high affinities for message passing. This future affinity learning and top-$k$ filtering strategy enables proximity in feature space to represent the potential relationships between future positions. Extensive experiments on the large-scale Argoverse motion forecasting dataset show that our approach captures the future interaction properly and leads to superior performance in multi-agent motion prediction.

%Our contributions are summarized as follows:
%\begin{itemize}
%	\setlength{\itemsep}{0ex}
%	\item We propose a novel motion prediction framework that models potential future interactions as well as history interactions in an end-to-end manner.
%	\item We present a future decoder that extracts potential future information, enabling to capture future interactions before computing output motions.
%	\item We show that interacting agent pairs can be fairly found by our affinity learning and top-$k$ filtering without explicit future state information.
%	\item We conduct extensive experiments on the large-scale Argoverse motion forecasting dataset while achieving superior performance. 
%\end{itemize}

\section{Related Work}
\label{sec:related}
%\subsection{Scene Context Encoding}The motion prediction problem requires the extraction of abundant information from a traffic scene, given the past motions of agents and a high-definition (HD) map. Early works~\cite{chai2019multipath, cui2019multimodal, gilles2021home, hong2019rules, salzmann2020trajectron++} adopted rasterization, which represents the input data as a bird-eye-view image to take advantage of off-the-shelf 2D CNNs. However, image representation causes inevitable information loss in quantization and its performance is considerably influenced by the spatial resolutions of the images. Recent works~\cite{zhou2022hivt, wang2022ltp, gilles2021thomas, gu2021densetnt, ngiam2021scene,liu2021multimodal} tend to adopt graph structure with vectorized data, following pioneering works such as VectorNet~\cite{gao2020vectornet} and LaneGCN~\cite{liang2020learning}. These approaches consider each entity as a set of vectorized polylines and learn the edge relation between entity nodes by applying GNNs. We follow this graph representation as it shows good scalability and effectiveness.

\begin{figure*}[t]
	\begin{center}
		\includegraphics[width=\linewidth]{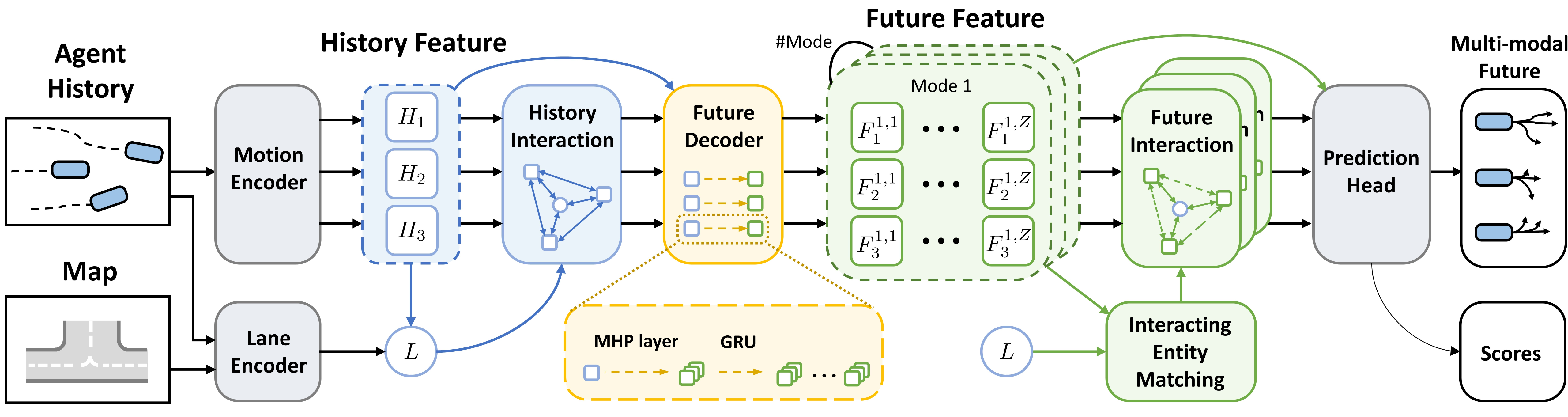}
		
	\end{center}
	\vspace{-0.6cm}
	\caption{Architecture of FIMP framework. Our network consists of two parts each for history and future feature learning. The future decoder separates the future feature space from the history, enabling the interaction modeling in respective time zones.
	}
	\vspace{-0.5cm}
	\label{fig:2}
\end{figure*}
In complex traffic scenarios as in urban areas, diverse interactions occur between multiple entities simultaneously. As the future behaviors of agents are considerably affected by neighboring entities, it is crucial to capture the interactions between them in spatial as well as temporal perspectives. Existing interaction modelings mainly fall into two approaches: observed-historical-information-based interaction and estimated-future-states-based conditional prediction.

\noindent\textbf{History based interaction.}\hspace{0.3cm}Numerous methods attempt to find the interaction by extracting useful information from observed data in history timesteps. LaneGCN~\cite{liang2020learning} models four types of interactions: actor-to-lane, lane-to-lane, lane-to-actor and actor-to-actor. These interactions are captured in series by propagating the spatial information over the lane graph. HiVT~\cite{zhou2022hivt} finds the interacting entity pairs in each past timestep using observed positions and applies scaled dot-product attention to learn the interactions. However, these interaction modelings based on history features extracted from observed data are not sufficient to purely model the future interaction, which considerably affects the future behaviors of agents. In contrast, our FIMP decouples future feature from history feature to represent potential future information aside from the observed data. By separating agent embedding into future and history, we can model the future interaction as well as history interaction in an end-to-end manner.

\noindent\textbf{Conditional prediction.}\hspace{0.3cm}To consider the future interaction explicitly, some methods~\cite{salzmann2020trajectron++, sun2022m2i, tolstaya2021identifying, khandelwal2020if, park2023leveraging} explore the conditional prediction that takes the future states of another agent as an input. CBP~\cite{tolstaya2021identifying} takes the ground truth future motion of the query agent and models the behavior changing of a target agent. M2I~\cite{sun2022m2i} learns to predict the relationships between agents by classifying them as pairs of influencer and reactor, and produces the reactor's trajectory conditioned on the estimated influencer's trajectory. The concurrent work FRM~\cite{park2023leveraging} models the future interaction by predicting the lane-level waypoint occupancy explicitly. Then the agents passing the adjacent lanes are regarded as an interacting pair. These approaches can consider potential interactions in the future, but they require access to the ground-truth motion or should explicitly predict the approximate motion of an influencing agent. Furthermore, the traffic scenarios with mutual interaction are disregarded in most methods. In contrast, our FIMP does not require the prior knowledge of high-level motion cues because implicit future information is used to capture the potential interaction at an intermediate feature-level. We demonstrate that interacting pairs and their connectivity in the future can be well-identified by our future affinity learning and top-$k$ filtering strategy.

\section{Method}
In this section, we first introduce the formulation of multi-head attention in Section~\ref{sec:mha} and subsequently elaborate on our approach. The architecture of our framework is illustrated in Figure~\ref{fig:2}.

\subsection{Multi-Head Attention}
\label{sec:mha}
We exploit multi-head attention (MHA) to model the interaction and temporal dependencies in our method. Following~\cite{vaswani2017attention}, we define an MHA with $h$ heads based on a scaled dot-product attention for input variables $X$ and $Y$:
\begin{equation} \label{eq:1}
\text{MHA}(X,Y) = [\text{Attn}_1(X,Y),...,\text{Attn}_h(X,Y)]W^O,
\end{equation}
\begin{equation} \label{eq:2}
\text{Attn}_i(X,Y) = \text{softmax}({(XW_i^Q)(YW_i^K)^\top \over \sqrt{C \slash h}})YW_i^V,
\end{equation}
where projections $W_i^Q, W_i^K, W_i^V \in \mathbb{R}^{C \times C\slash h}$ and $W^O \in \mathbb{R}^{C \times C}$ are parameter matrices, $C$ is the feature dimension and $[\cdot,\cdot]$ indicates concatenation. The multi-head self-attention (MHSA) can be represented as MHA with two identical inputs:
\begin{equation} \label{eq:3}
\text{MHSA}(X) = \text{MHA}(X,X).
\end{equation}

\subsection{Input Representation}
For input data, we adopt a vectorized representation that involves geometric attributes of entities as a form of vector sets. As the raw vectorized data is not invariant to translation and rotation, we transform the absolute history positions of each agent to be agent-centric where the scene is centered at the current position of a target agent and aligned with its heading. Specifically, we denote the motion history of an agent $i$ as $A_i = \left\lbrace \Delta p_i^t  \right\rbrace_{t=1}^{T_{history}} $, where $\Delta p_i^t \in \mathbb{R}^2$ is a 2D motion vector from timestep $t-1$ to $t$ and $T_{history}$ is the number of history timesteps. Then the multi-agent history can be denoted as
$A_{input} = \left\lbrace A_i \right\rbrace_{i=1}^N \in \mathbb{R}^{N \times T_{history} \times 2}$, where $N$ is the number of agents. To capture the interactions, we sample the neighboring agents for each target agent in the current frame $t=T^{history}$ within a predefined radius. The relative position and heading of sampled agent $j$ in the coordinates centered at target agent $i$ are represented as $\left\lbrace p_{ij}, \theta_{ij}\right\rbrace$.  We also sample the set of lane segments that are close to the current position of each agent and convert them into corresponding agent-centric coordinates. We represent the lane segment $\omega$ surrounding agent $i$ as $\left\lbrace l_{i\omega}^{start}, l_{i\omega}^{end}\right\rbrace$, where $l_{i\omega}^{start}$ and  $l_{i\omega}^{end}$ are starting and ending positions of the lane segment in the coordinates centered at agent $i$.

\subsection{History Feature Encoding}
\noindent\textbf{Motion encoder.}\hspace{0.3cm}Given the motion history $A_{input} \in \mathbb{R}^{N \times T_{history} \times 2}$, we first encode the motion vector at each timestep by using Multi-Layer Perceptrons (MLPs) to obtain the motion embedding $A_m \in \mathbb{R}^{N \times T_{history} \times C}$:
\begin{equation} \label{eq:4}
A_{m} = \text{MLP}(A_{input}).
\end{equation}
Then, we learn the temporal dependencies across the history sequence of $A_m$ by adopting a temporal attention module. Similar to ViT~\cite{dosovitskiy2020image}, each layer in the module is comprised of layer norm (LN) operations~\cite{ba2016layer}, MHSA block, residual connections~\cite{he2016deep}, and a feed-forward network (MLP). A learnable token with size $\mathbb{R}^{C}$ is added to the end of input sequence, which reasons the sequence of motions as a whole. Taking $A_m$ as an input for the initial layer, the output $A_m^{l+1}$ from the $l^{th}$ encoder layer is obtained as
\begin{equation} \label{eq:5}
\hat{A}_m^l = \text{MHSA}(\text{LN}(A_m^l+p)) + (A_m^l+p),
\end{equation}
\begin{equation} \label{eq:6}
A_m^{l+1} = \text{MLP}(\text{LN}(\hat{A}_m^l)) + \hat{A}_m^l,
\end{equation}
where positional embeddings $p \in \mathbb{R}^{(T_{history}+1) \times C}$ is added to the motion embedding. Finally, we adopt the updated learnable token in the last motion embedding as the history feature $H \in \mathbb{R}^{N \times C}$, which represents the motion history information of $N$ agents.

\noindent\textbf{History interaction.}\hspace{0.3cm}We extract the agent-map and agent-agent interactions in the history to better understand the observed scene. %For each target agent, the neighboring lane segments and agents are sampled within the predefined local radius.
%We first sample the neighboring agents and lane segments of each interested agent within the predefined local radius. The vectors of sampled entities are then translated and rotated into corresponding agent-centric coordinate. 
For target agent $i$, we first encode the surrounding lane $\omega$ in the agent-centric coordinates by using MLPs to obtain the lane embedding $L_{i\omega} \in \mathbb{R}^{C}$:

\begin{equation} \label{eq:7}
L_{i\omega} = \text{MLP}([l_{i\omega}^{start},\hspace{0.1cm} l_{i\omega}^{end} - l_{i\omega}^{start}, attr_\omega]),
\end{equation}
where $attr_\omega$ is the lane attributes (e.g., turning direction, lane type). The obtained lane embedding $L_{i\omega}$ is then incorporated into target agent $i$ by MHA to obtain lane-aware agent feature $H_i^L$:
\begin{equation} \label{eq:8}
H_i^{L} = \text{MHA}(H_i, \left\lbrace L_{i\omega} \mid \omega \in N_L(i) \right\rbrace),
\end{equation}
where $N_L(i)$ is the set of neighboring lanes respective to agent $i$. To capture the agent-agent interaction, we first project the feature $H_j$ of neighboring agent $j$ into the coordinate of target agent $i$ by encoding the relation between their agent-centric coordinates:
\begin{equation} \label{eq:9}
H_{ij}^L = \text{MLP}(H_j^L) + \text{MLP}([p_{ij},cos(\theta_{ij}), sin(\theta_{ij})]).
\end{equation}
The projected feature $H_{ij}^L$ can be learned by the relative position $p_{ij}$ and heading $\theta_{ij}$. Then we incorporate the neighboring agents' feature by MHA to obtain interaction-aware history feature $\tilde{H}_i$:

\begin{equation} \label{eq:10}
\tilde{H}_i = \text{MHA}(H_i^L, \left\lbrace H_{ij}^L \mid j \in N_A(i), \right\rbrace),
\end{equation}
where $N_A(i)$ is the set of neighboring agents respective to agent $i$.

\begin{figure*}[t]
	\begin{center}
		\includegraphics[width=\linewidth]{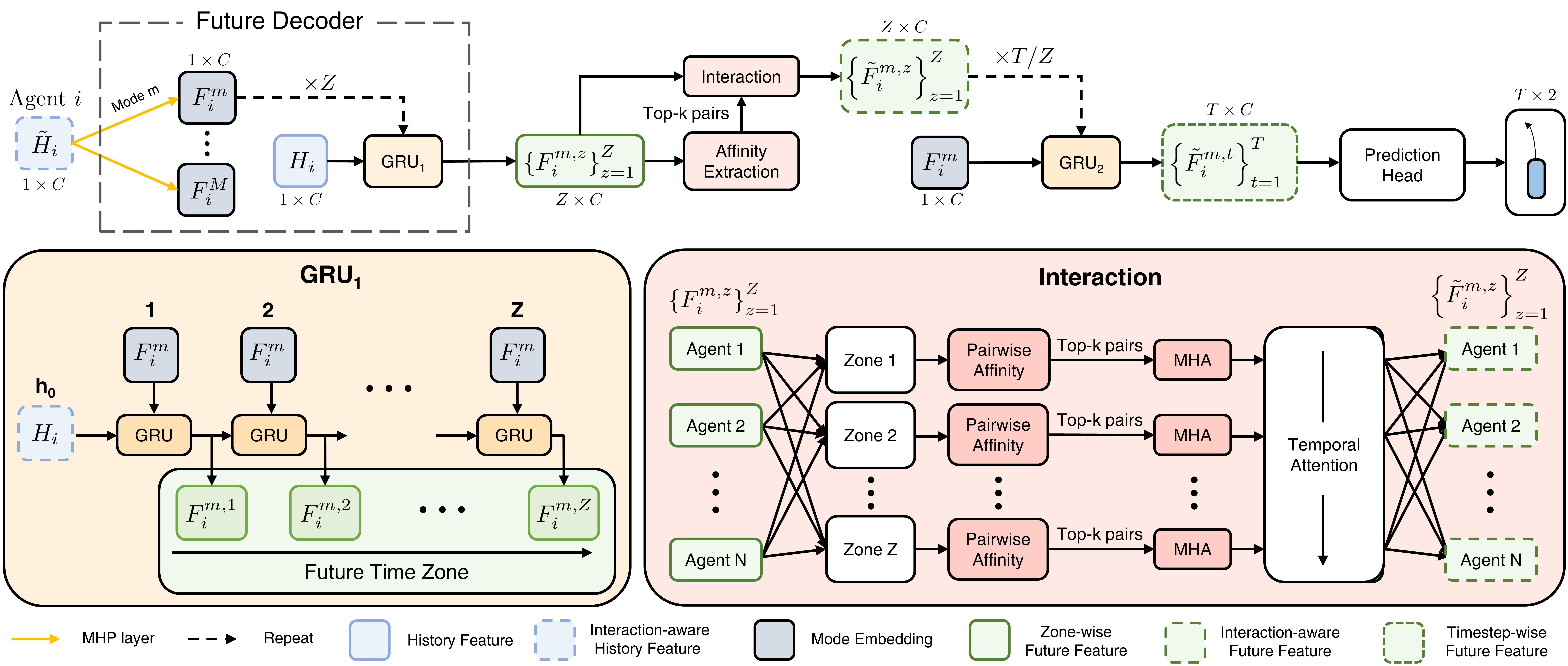}
		
	\end{center}
	\vspace{-0.6cm}
	\caption{Network structure from future decoder to prediction head. For brevity, the process to predict the motions on a single mode of agent $i$ is illustrated. $\text{GRU}_2$ works similar to $\text{GRU}_1$, but interaction-aware zone-wise future feature $\tilde{F}^{m,z}$ is only repeated for timesteps that it involves. 
	}
	\vspace{-0.5cm}
	
	\label{fig:3}
\end{figure*}

\subsection{Future Decoder}Our approach adopts an intermediate future decoder that decouples the future feature from history feature. It aims to derive features that represent the potential future information aside from historical information, enabling feature-level future interaction modeling before predicting motions. The decoder comprises two primary components: a multi-head projection (MHP) layer for multi-modal prediction and GRU for temporalization. The MHP layer generates diverse future mode embeddings from the history feature $\tilde{H}$ using different MLPs for each mode. The $m^{th}$ mode embedding $F^m \in \mathbb{R}^{N \times C}$ can be computed by $\text{MLP}_m$ as
\begin{equation} \label{eq:11}
F^m = \text{MLP}_m(\tilde{H}), \hspace{0.2cm} m \in \left\lbrace 1,2,...,M \right\rbrace,
\end{equation}
where $M$ is the number of modes. Then GRU is used to temporalize the mode embedding into several \textit{future time zones}, with each zone containing a predefined number of timesteps. When the number of future timesteps to be predicted and time zones are $T$ and $Z$ respectively, the number of future timesteps in each zone becomes $T \slash Z$. We divide the future period into these sparse time zones instead of dense timesteps because it is more effective to map uncertain future information, and the interaction tends to occur over consecutive timesteps. As illustrated in Figure~\ref{fig:3}, a GRU takes a history feature $H$ obtained from the motion encoder as an initial hidden state $h_0$ and uses the identical mode embedding $F^m$ as the inputs for all time zones, different from conventional GRU that takes temporal inputs. By repeating non-temporal mode embedding as a input sequence, we can temporalize the embedding into zone-wise future feature $F^{m,z} \in \mathbb{R}^{N \times C}$, where $z \in \left\lbrace 1,2,...,Z \right\rbrace$ is an index of time zones. 

\noindent\textbf{Learning.}\hspace{0.3cm}After interactions are captured within respective zones of future features, another GRU is used to temporalize the zone into timesteps. Similar to GRU in the future decoder, mode embedding is used as the initial hidden state and future features of each zone become the inputs for timesteps that it covers. The obtained timestep-wise future feature $\tilde{F}^{m,t}$ with $t \in \left\lbrace 1,2,...,T \right\rbrace$ is then used to predict the position at the corresponding timestep. In this learning, future feature of each time zone is optimized to represent the potential states in a particular period over the involved future timesteps.

%\begin{equation} \label{eq:1}
%h_t = \text{GRU}(h_{t-1}, A^m_{mode}),
%\end{equation}
\subsection{Future Interaction Modeling}
In this section, we describe our approach for identifying and modeling potential future interactions among entities.

\noindent\textbf{Agent-lane interaction.}\hspace{0.3cm}As the map topology is time-invariant, agent-lane interaction is less difficult to capture than agent-agent interaction. We thus roughly incorporate the lane information within the potential future trajectories of agents, similar to history interaction modeling. We borrow the lane embedding $l_{i\omega}$ computed in Eq.~\ref{eq:7} and apply MHA to future feature $F^m$ as in Eq.~\ref{eq:8}, but with varied sampling regions to cover the lanes around future motions. The region is defined roughly with a bigger radius considering the heading of an agent, and the attention module learns to selectively incorporate the interacting lanes.

\noindent\textbf{Agent-agent interaction via affinity learning.}\hspace{0.3cm}
We identify the interacting agent pairs by learning the affinity between implicit future features. It aims to extract high affinity from agent pairs where message passing is required owing to potential interaction. As the encoded information in $F^{m,z}$ is based on each agent-centric coordinate, it is necessary to transform the future features of all agents into the same feature space before computing affinity. We therefore project them into an autonomous vehicle (AV)'s coordinates as the reference feature space. Similar to Eq.~\ref{eq:9} in history interaction modeling,
the projected future feature $F_{\alpha i}^{m,z} \in \mathbb{R}^{C}$ of agent $i$ in the coordinate of AV $\alpha$ can be obtained as
\begin{equation} \label{eq:12}
F_{\alpha i}^{m,z} = \text{MLP}(F_i^{m,z}) + \text{MLP}([p_{\alpha i},cos(\theta_{\alpha i}), sin(\theta_{\alpha i})]).
\end{equation} 
Here, the affinity matrix between projected future features $F_{\alpha i}^{m,z}$ is computed based on feature distance and used to determine which agent pair to model the future interaction from. For each target agent, only the agents with top-$k$ high affinities are selected as the interacting pairs and can perform the message passing. The interaction-aware future feature $\tilde{F}_i^{m,z}$ is obtained by MHA as follows:
\begin{equation} \label{eq:13}
\tilde{F}_i^{m,z} = \text{MHA}(F_i^{m,z}, \left\lbrace F_{ij}^{m,z} \mid j \in \text{Top}^{m,z}_k(i) \right\rbrace),
\end{equation} 
where $\text{Top}^{m,z}_k(i)$ indicates the indices of interacting agents. 
In this process, $F_{\alpha i}^{m,z}$ is learned to represent the future states of agents in the reference coordinates so that the relationships between agents' future positions can be indicated by the proximity in reference feature space $\mathbb{R}^C$. The optimization of affinity learning enables interacting agents to perform message passing with high affinities, leading to better prediction. This top-$k$ filtering strategy is ideal for interaction modeling as only a few agents will be in interaction while others are noises. After learning, we empirically verify in Section~\ref{sec:ablation} that our affinity learning with top-$k$ filtering enables proper identification of interacting agents in future timesteps, compared to conventional interacting agents matching strategies of existing methods.

\begin{comment}
\begin{equation} \label{eq:1}
S_{ij} = (F_iW^Q)(F_{ij}W^K)^\top,
\end{equation}
\begin{equation} \label{eq:1}
\alpha_{ij} = {\text{exp}(S_{ij}) \over \sum_{q \in \text{Top}_i^k} \text{exp}(S_{iq})}, \text{if} \hspace{0.1cm} j \in \text{Top}_i^k,
\end{equation}
\begin{equation} \label{eq:1}
m_i = \sum_{j \in \text{Top}^k_i} \alpha_{ij}(F_{ij}W^V)
\end{equation}
\end{comment}

As the interaction is considered at each time zone independently, we further capture the temporal information by applying temporal attention module in the same way as described in Eqs.~\ref{eq:5} and \ref{eq:6}.

\subsection{Multi-Modal Motion Prediction}
\noindent\textbf{Multi-modality.}\hspace{0.3cm}As the future behaviors of agents are not deterministic, the network should predict the multi-modal motions to deal with environmental uncertainties. Our FIMP thus generates multiple mode embeddings in the future decoder and extracts timestep-wise future features $\tilde{F}^{m,t}$ for each mode. An MLP in the prediction head takes these features to output the final forecasting as a form of Laplace distribution with location $\mu_i^t \in \mathbb{R}^2$ and scale $b_i^t \in \mathbb{R}^2$, which represent the state of an agent at future timestep $t$. Following~\cite{ye2021tpcn}, the displacement error $d^m_i$ at the endpoint is predicted for each mode and is converted to the confidence score with softmin function.

\noindent\textbf{Training objective.}\hspace{0.3cm}We train our model end-to-end with the regression loss $\mathcal{L}_{reg}$ and the classification loss $\mathcal{L}_{cls}$. The winner-takes-all (WTA) strategy is used in regression tasks to avoid penalizing diverse plausible predictions. The best prediction ($\hat{\mu}_i^t, \hat{b}_i^t$) among $M$ modes is selected for each agent by calculating the average displacement error along timesteps. Then we use the negative log-likelihood as
\begin{equation} \label{eq:14}
\mathcal{L}_{reg} = -{1 \over {NT}}\sum_{i=1}^{N}\sum_{t=1}^{T}logP(g_i^t \mid \hat{\mu}_i^t, \hat{b}_i^t),
\end{equation}
where $g_i^t$ is the ground-truth location of agent $i$ at future timestep $t$ and $P(\cdot\hspace{-0.2cm}\mid\hspace{-0.2cm}\cdot)$ is a probability density function of Laplace. For classification loss $\mathcal{L}_{cls}$, we adopt smooth $L_1$ loss between ground-truth displacement and prediction. The final loss is the sum of regression and classification losses with equal weights.

\section{Experiments}
\subsection{Experimental Setup}
\noindent\textbf{Dataset.}\hspace{0.3cm}We evaluate our approach on the large-scale Argoverse motion forecasting dataset~\cite{chang2019argoverse}, which comprises approximately 320K real-word driving scenarios. Each data comprises the trajectories of agents and an HD map. The training and validation set contains 5-second scenarios sampled at 10 Hz, where the first 2-second trajectories are given as an input and remaining 3-seconds are labeled as future trajectories to be predicted. For the test set, only the first 2-second observations are given.

\noindent\textbf{Metrics.}\hspace{0.3cm}As motion prediction is multi-modal by nature, we adopt the widely used metrics for multi-modal evaluation, including minimum Final Displacement Error (minFDE) and minimum Average Displacement Error (minADE). minFDE is the L2 distance between the best predicted trajectory and ground-truth trajectory at the final future timestep, while minADE is the error averaged over all future timesteps. Argoverse benchmark allows up to $M=6$ predictions for each agent and we predict six trajectories following the baseline.

\noindent\textbf{Implementation details.}\hspace{0.3cm}We train our model in an end-to-end manner by using an AdamW~\cite{loshchilov2017decoupled} optimizer for 64 epochs with two Titan RTX GPUs. We use a batch size of 32 and an initial learning rate of 0.0005, which decays with a cosine annealing scheduler~\cite{loshchilov2016sgdr}. The agent-lane and agent-agent interaction modules are comprised of 1 and 3 layers respectively, while the temporal attention module consists of 4 layers. Each MHA block has 8 heads and the feature dimension $C$ is 128. The local radii for neighboring lane and agent sampling used in history interactions are both 50m, and the radius for neighboring lane sampling used in future interaction is 100m.

\begin{table}[!h]
	\caption{Results on the Argoverse validation and test set.}
	\vspace{-0.8cm}
	\begin{center}
		\resizebox{\linewidth}{!}{
			\begin{tabular}{l|cc|cc}
				\toprule
				\multirow{2}{*}{Model} &   \multicolumn{2}{c|}{Validation set} & \multicolumn{2}{c}{Test set}  \\
				& minFDE & minADE & minFDE & minADE \\
				\midrule
				LaneGCN \cite{liang2020learning}  &1.08  & 0.71&1.36  & 0.87  \\
				mmTrans \cite{liu2021multimodal}& 1.15 & 0.71 & 1.34 & 0.84\\
				TPCN \cite{ye2021tpcn} & 1.15  & 0.73  &1.24  & 0.82 \\
				DenseTNT \cite{gu2021densetnt} & 1.05 & 0.73  & 1.28 & 0.88   \\
				GOHOME \cite{gilles2022gohome} & 1.26 & -  & 1.45 & 0.94  \\
				PAGA \cite{da2022path}& 1.02  &  0.69 & 1.21  & 0.80  \\
				THOMAS \cite{gilles2021thomas} & 1.22 & - & 1.44 & 0.94  \\
				AutoBot \cite{girgis2021latent} & 1.10 & 0.73   & 1.37 & 0.88 \\
				Scene Transformer \cite{ngiam2021scene} &  - & -   &  1.23 & 0.80   \\
				LTP \cite{wang2022ltp} & 1.07  &  0.78& 1.29  &  0.83 \\
				HiVT \cite{zhou2022hivt}  & 0.96  &  0.66 & 1.17  &  0.77 \\
				FRM \cite{park2023leveraging}  & 0.99  &  0.68 & 1.27  &  0.82 \\
				%			HoliGraph & -  &  - & 1.65 & 0.98\\
				%			GANet & 0.93  &  0.67 & 1.16 & 0.80\\
				\hline
				FIMP (ours) &  $\textbf{0.92}$ & $\textbf{0.64}$  & $\textbf{1.13}$ & $\textbf{0.76}$   \\
%				FIMP w/o Future Interaction&  0.97 & 0.67   & 1.26 & 0.82   \\
				\bottomrule
		\end{tabular}}
		\vspace{-0.9cm}
		\label{tab:argo}
	\end{center}
\end{table}

\subsection{Evaluation}
\label{eval}

\noindent\textbf{Quantitative results.}\hspace{0.3cm}We compare our FIMP with state-of-the-art methods that have recently been applied on the Argoverse dataset. The results on validation and test sets are reported chronologically in Table~\ref{tab:argo}. FIMP achieves the best performance on the validation set by a clear margin in terms of minFDE and minADE. The results on the test set also outperform the related works. For all metrics on both sets, our future interaction modeling improves the performance remarkably, reducing the large proportions of prediction errors. As our future interaction modeling is compatible with other state-of-the-art methods, we believe that the result can be further improved by adopting a stronger baseline.

Among considered studies, the latest concurrent method FRM~\cite{park2023leveraging} also aims to capture the future interaction by explicitly predicting the lane-level waypoint occupancy first and subsequently performing conditional prediction, but shows inferior performance. This is because the occupancy along lane axis is predicted without considering possible interactions, and the final forecasting is heavily reliant upon the accuracy of this occupancy estimation. Furthermore, the approach dependent on lane topology cannot be used in the map where lane information is not available or poorly constructed, and 2-phase motion prediction inference is not suitable for real-time applications. In contrast, our FIMP extracts implicit future information before computing motions in an end-to-end manner, which is optimized to determine when and where agents will be in the future. Based on our experiments, it appears that using this intermediate feature-level information to represent potential future states is adequate for identifying interacting agents and their connectivity, without encountering issues with pre-estimation errors. 

\noindent\textbf{Qualitative results.}\hspace{0.3cm}We present the qualitative results of FIMP compared to related models in Figure~\ref{fig:4}. For LaneGCN~\cite{liang2020learning} and HiVT~\cite{zhou2022hivt}, we use the pretrained models provided by authors. We can observe that our FIMP captures the potential interaction in the future and predicts the plausible trajectories for both agents without any conflicts. In contrast, other historical-information-based methods lack capability of capturing the future motions of an interacting agent, resulting in unrealistic trajectory overlaps. We provide more comparison and analysis on diverse traffic scenarios in the supplementary material.

\begin{figure}[t]
	\begin{center}
		\includegraphics[width=\linewidth]{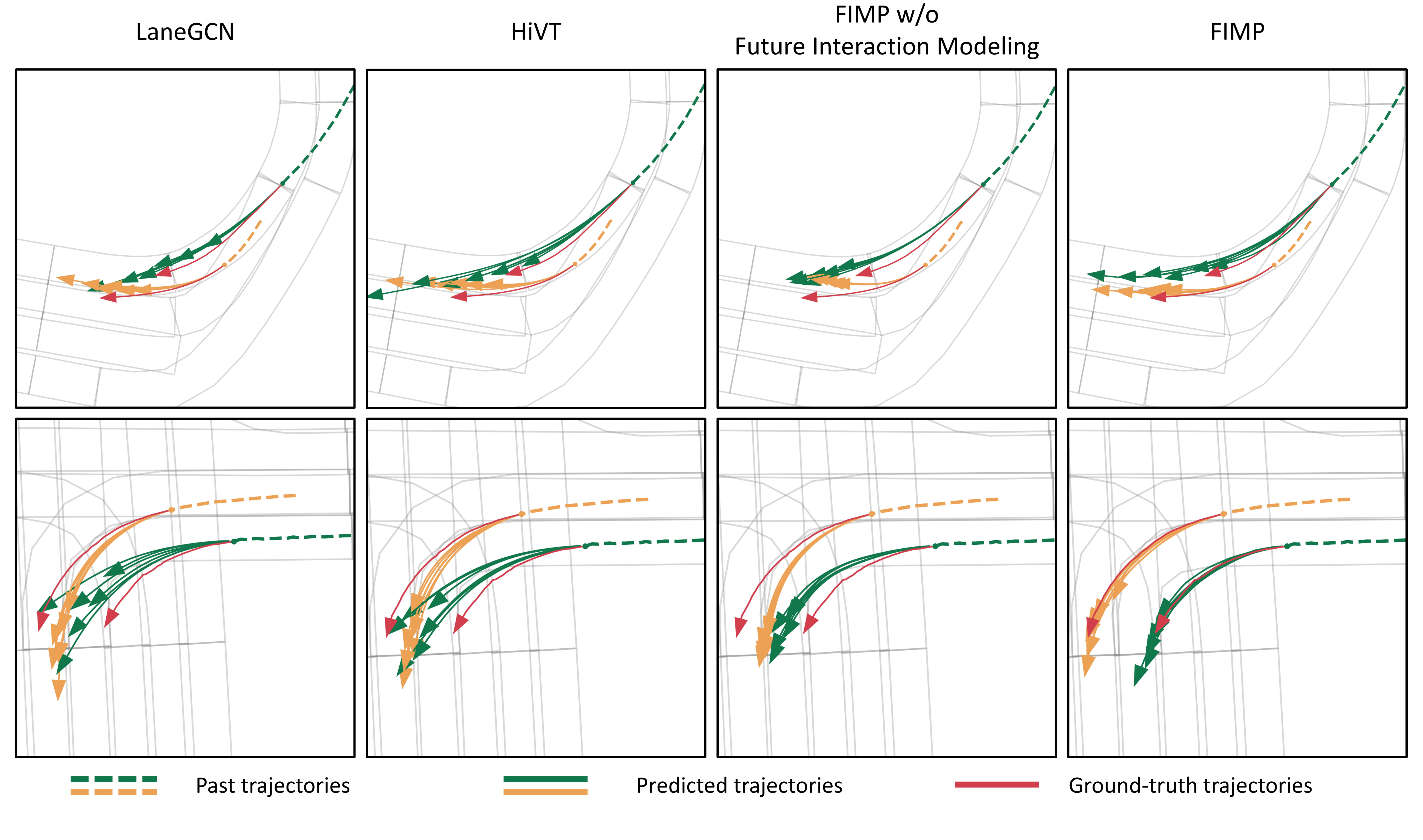}
		
	\end{center}
	\vspace{-0.8cm}
	\caption{Qualitative comparison of models in the scenarios where future interaction modeling is essential. The trajectories of interacting agents are shown in green and orange while ground-truth trajectories are in red.
	}
	\vspace{-0.1cm}
	\label{fig:4}
\end{figure}

\subsection{Ablation Study}
\label{sec:ablation}
We conduct an ablation study to demonstrate the effectiveness of each component in FIMP. All ablation results are based on the Argoverse validation set.

\noindent\textbf{Importance of future interaction.}\hspace{0.3cm}We present the contributions of history and future interaction modelings in Table~\ref{tab:interaction}. The first two rows show that capturing the interactions in agent-lane and agent-agent pairs during the observed timesteps can considerably improve the performance of motion prediction. This finding is consistent with prior research in this field. However, a comparison between rows 2 and 3 reveals that it is much more significant to consider the potential future interactions to predict the agents' future motions, and our proposed FIMP captures such interactions effectively. It is also observed in row 4 that the agent-agent interaction in the history timesteps becomes less significant when the network is able to model the future interaction. Furthermore, incorporating agent-lane future interaction modeling can enhance the performance of minADE as presented in row 5.
\begin{table}[t]
	\caption{Ablation study on interaction modeling.}
	\vspace{-0.5cm}
	\begin{center}
		\resizebox{0.8\linewidth}{!}{
			\begin{tabular}{c|cc|cc|cc}
				\toprule
				& \multicolumn{2}{c|}{History} &  \multicolumn{2}{c|}{Future} &  \multirow{2}{*}{minFDE} & \multirow{2}{*}{minADE}     \\
				&A-L & A-A & A-L & A-A & &  \\
				\midrule
				%A && &&   & 1.23 & 0.79 & 0.18\\
				1&\ding{51}& &&   & 1.17 & 0.74 \\
				2&\ding{51}& \ding{51}&&  &  0.97 &  0.67 \\
				3&\ding{51}& & & \ding{51}  & 0.93  &0.65  \\
				4&\ding{51} & \ding{51}& & \ding{51}  & \textbf{0.92}  &0.65  \\
				5&\ding{51} &\ding{51}&\ding{51}&\ding{51}  &  \textbf{0.92} &  \textbf{0.64} \\
				\bottomrule
		\end{tabular}}
		\label{tab:interaction}
	\end{center}
	\vspace{-0.7cm}
\end{table}

\begin{table}[t]
	\caption{Ablation study on strategies of matching interacting agents in the future.}
	\vspace{-0.5cm}
	\begin{center}
		\resizebox{0.8\linewidth}{!}{
			\begin{tabular}{cc|cc}
				\toprule
				Interacting agent & top-$k$ & minFDE & minADE \\
				\midrule
				%All   & \ding{55}   &  &  & \\
				Local region (r=50)  & \ding{55}   & 0.96 & 0.66 \\
				Local region (r=100)  & \ding{55}   & 0.95 & 0.66\\
				Nearest order & 5 & 0.96 & 0.67 \\
				Nearest order & 10 &  0.96 & 0.66 \\
				High future affinity   & 5   &  0.93 &  0.65\\
				High future affinity   & 10   &  \textbf{0.92} &  \textbf{0.64}\\
				High future affinity   & 20 &  0.93 &  \textbf{0.64}\\
				\bottomrule
		\end{tabular}}
		\label{tab:pair}
	\end{center}
	\vspace{-0.2cm}
\end{table}

\begin{table}[t]
	\caption{Ablation study on the number of future time zones and inference latency.}
	\vspace{-0.5cm}
	\begin{center}
		%		\resizebox{\linewidth}{!}{
		\begin{tabular}{cc|cc|c}
			\toprule
			$Z$  & \#timesteps & minFDE & minADE &  Latency (ms/agent)  \\
			\midrule
%			3    & 10 & 0.94 & 0.65  & 161 \\
%			5      & 6 & \textbf{0.92} &  \textbf{0.64}& 220 \\
%			6    &  5 &\textbf{0.92} &  0.65& 254  \\
%			10    &  3 & 0.94 &  0.65& 342 \\
			3    & 10 & 0.94 & 0.65  & 17 \\
			5      & 6 & \textbf{0.92} &  \textbf{0.64}& 24 \\
			6    &  5 &\textbf{0.92} &  0.65& 28  \\
			10    &  3 & 0.94 &  0.65& 37 \\
			\bottomrule
		\end{tabular}
		\label{tab:timezone}
	\end{center}
	\vspace{-0.7cm}
\end{table}

\noindent\textbf{Interacting pair matching strategy.}\hspace{0.3cm}In Table~\ref{tab:pair}, we investigate three distinct methods for discerning interacting agents in future timesteps: local-region-based matching, nearest-order-based matching, and high-future-affinity-based matching. The first two methods operate on the assumption that agent pairs satisfying the matching criteria at the current time step will interact in the future, akin to historical interaction modeling. However, the results reveal that these conventional methods struggle to accurately identify interacting pairs in the future, as the relationships between agents in the future do not consistently align with those at the current timestep. In contrast, our future-affinity-based matching converge to better results by identifying the top-$k$ interacting agents in the future properly. The choice of $k$ is not too sensitive but $k=10$ leads to best performance. 

\noindent\textbf{Number of future time zones $Z$ and latency.}\hspace{0.3cm}As it is ineffective to consider the interaction across all future timesteps, our intermediate future decoder derives information for sparse time zones instead of dense timesteps. We divide the $T=30$ future timesteps into $Z$ time zones, capturing future interaction within each respective zone. To evaluate the impact of the number of time zones, we conduct an ablation study by varying $Z$ as shown in Table~\ref{tab:timezone}. The results indicate that setting $Z=3$, with 10 timesteps in each zone, leads to worse performance compared to $Z=5$ or $6$ due to an aggregation of too many timesteps into a single zone. This aggregation makes it challenging to extract precise position information for the final prediction head. Conversely, setting $Z=10$ also results in poor performance, as the interaction is considered too densely along timesteps, allowing for redundant message passing. In addition, we measure the inference latency on the validation set using a batch size of 1 and a Titan RTX GPU. As our approach can make predictions for all agents simultaneously using a single forward, FIMP with $Z=5$ successfully predicts 6 possible trajectories of an agent at an average speed of 24ms while also capturing potential future interactions with other agents.

\section{Conclusion}
In this paper, we present a novel future interaction modeling for multi-agent motion prediction. Our model FIMP captures the potential interaction in an end-to-end manner by adopting a future decoder that derives the implicit future information aside from observed data. Experiments demonstrate that the interacting agent pairs and their relationships in the future can be effectively identified by learning future affinities and using top-$k$ filtering strategy. FIMP achieves superior performance on the Argoverse motion forecasting dataset with real-time inference, and future work can combine our future interaction modeling with other state-of-the-art methods to further build a strong framework.

\section{Additional Implementation Details}

\noindent\textbf{Lane preprocessing.}\hspace{0.3cm}The Argoverse dataset provides lane data, where each lane is represented by 10 points. From each lane, 9 lane segments are extracted as vectors between consecutive points. The lane segments surrounding agents are sampled and transformed into agent-centric coordinates by translation and rotation. Additionally, lane attributes are computed for each lane segment, including whether it is at an intersection, whether it has traffic control and whether it is from a left-turn lane or a right-turn lane.

%\medskip
%\noindent\textbf{Architecture details.}\hspace{0.3cm}
\medskip
\noindent\textbf{Efficient affinity extraction.}\hspace{0.3cm}We compute the affinity between projected future features based on negative L2 distance. As the na\"ive implementation of L2 distance leads to slow inference time, we simplify the computing process by a decomposition, as noted in \cite{kim2021lipschitz}:
\begin{equation} \label{eq:12}
	\begin{split}
		\text{Affinity}_{ij}^{m,z} & =-\Big\| F_{\alpha i}^{m,z} - F_{\alpha j}^{m,z} \Big\|^2_2 \\ & = 2F_{\alpha i}^{m,z}\cdot F_{\alpha j}^{m,z} - \Big\| F_{\alpha i}^{m,z} \Big\|^2_2 - \Big\| F_{\alpha j}^{m,z} \Big\|^2_2,
	\end{split}
\end{equation} 
where $\text{Affinity}_{ij}^{m,z}$ is the future affinity between agents $i$ and $j$ at mode $m$ and future time zone $z$. By selecting only top-$k$ agent pairs with high affinities, we can also reduce the number of expensive exponential function calls in softmax when capturing agent-agent interaction by multi-head attention.

\medskip
\noindent\textbf{Motion prediction with Laplace distribution.}\hspace{0.3cm}Our prediction head takes timestep-wise future features $\tilde{F}^{m,t}$ to output the final forecasting as a form of Laplace distribution with location and scale parameters. The activation function for scale parameter is $\text{ELU}(\cdot)+1+\epsilon$ where $\text{ELU}(\cdot$) is the exponential linear unit function and $\epsilon$ is set to 0.001.

\medskip
\noindent\textbf{Training details.}\hspace{0.3cm}We provide the model and training hyperparameters in Table~\ref{tab:1}. Our model is trained with two Titan RTX GPUs. We do not use any tricks such as ensemble methods and data augmentation.

\begin{table}[!t]
	\caption{Model and training hyperparameters.}
	\vspace{-0.8cm}
	\begin{center}
		\resizebox{\linewidth}{!}{
			\begin{tabular}{c|c}
				\toprule
				Hyperparameter & Value  \\
				\midrule
				Feature channel $C$& 128 \\
				\# Heads in a multi-head attention block & 8\\
				\# Agent-lane interaction layers& 1\\
				\# Agent-agent interaction layers& 3\\
				\# Temporal attention layers& 4\\
				Local radius for agent sampling (history)& 50m\\ 
				Local radius for lane sampling (history)& 50m\\
				Local radius for lane sampling (future)& 100m\\ 
				Top-$k$ affinities filtering & 10\\
				\# Future time zones $Z$ & 5\\
				Optimizer& AdamW\\ 
				Scheduler & Cosine annealing\\ 
				Initial learning rate& 0.0005 \\
				Weight decay & 0.0001\\
				Dropout & 0.1\\
				Batch size& 32\\
				Training epochs &64 \\
				\bottomrule
		\end{tabular}}
		\label{tab:1}
	\end{center}
\end{table}

\section{Additional Qualitative Comparisons}
\begin{figure*}[t]
	\begin{center}
		\includegraphics[width=\linewidth]{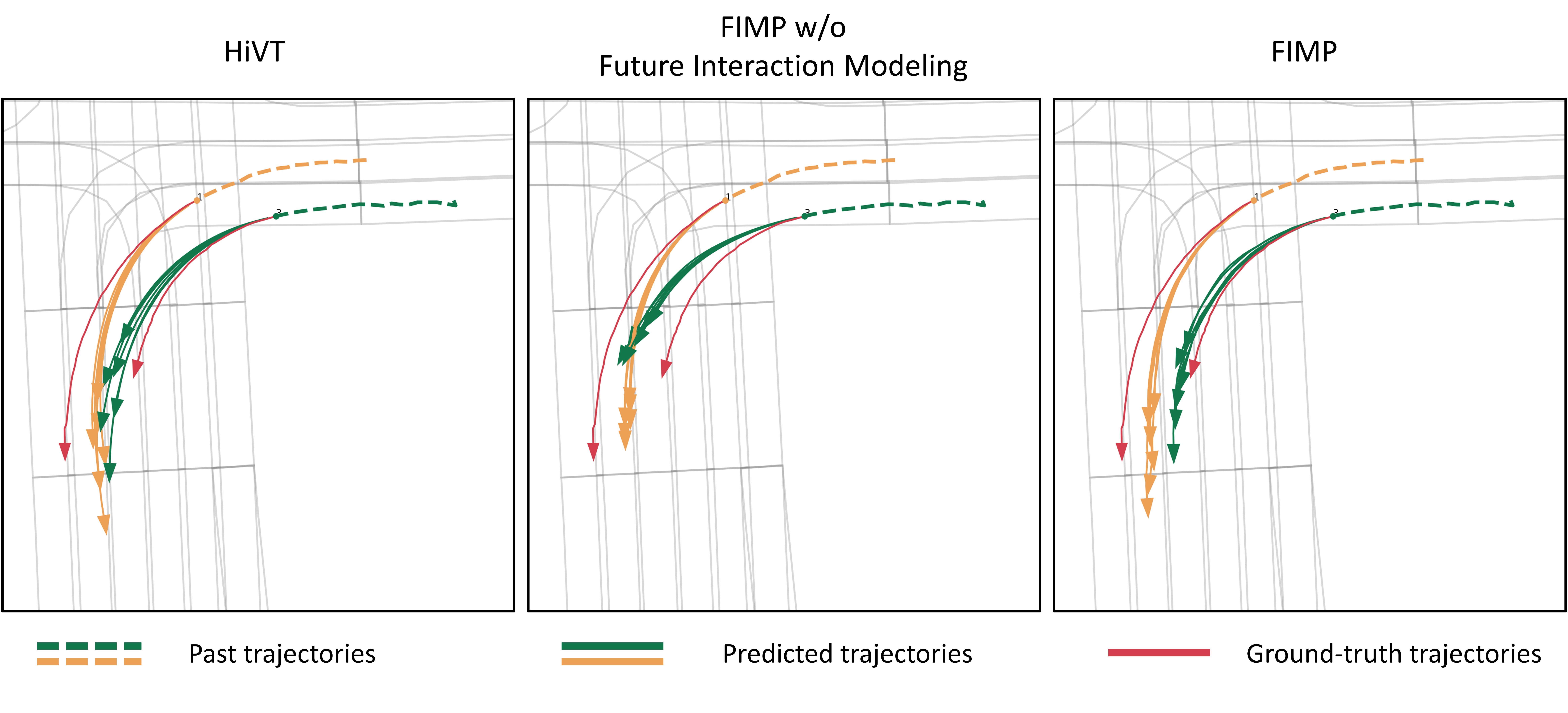}
		
	\end{center}
	\vspace{-0.5cm}
	\caption{In this scenario, the orange agent turns left in a wide curve, which is an unusual behavior, and all models fail to predict it. However, FIMP is still able to accurately predict the future motion of the green agent due to its ability to capture the future interactions between agents, whereas other models make incorrect predictions.
	}
	
	\label{fig:6}
\end{figure*}

From Figure~\ref{fig:1} to Figure~\ref{fig:6}, we present various prediction examples of FIMP on Argoverse validation set in comparison with our model without interaction modeling and state-of-the-art method HiVT~\cite{zhou2022hivt}. Only two interacting agents are visualized in each scene for clarity.

\begin{figure*}[t]
	\begin{center}
		\includegraphics[width=\linewidth]{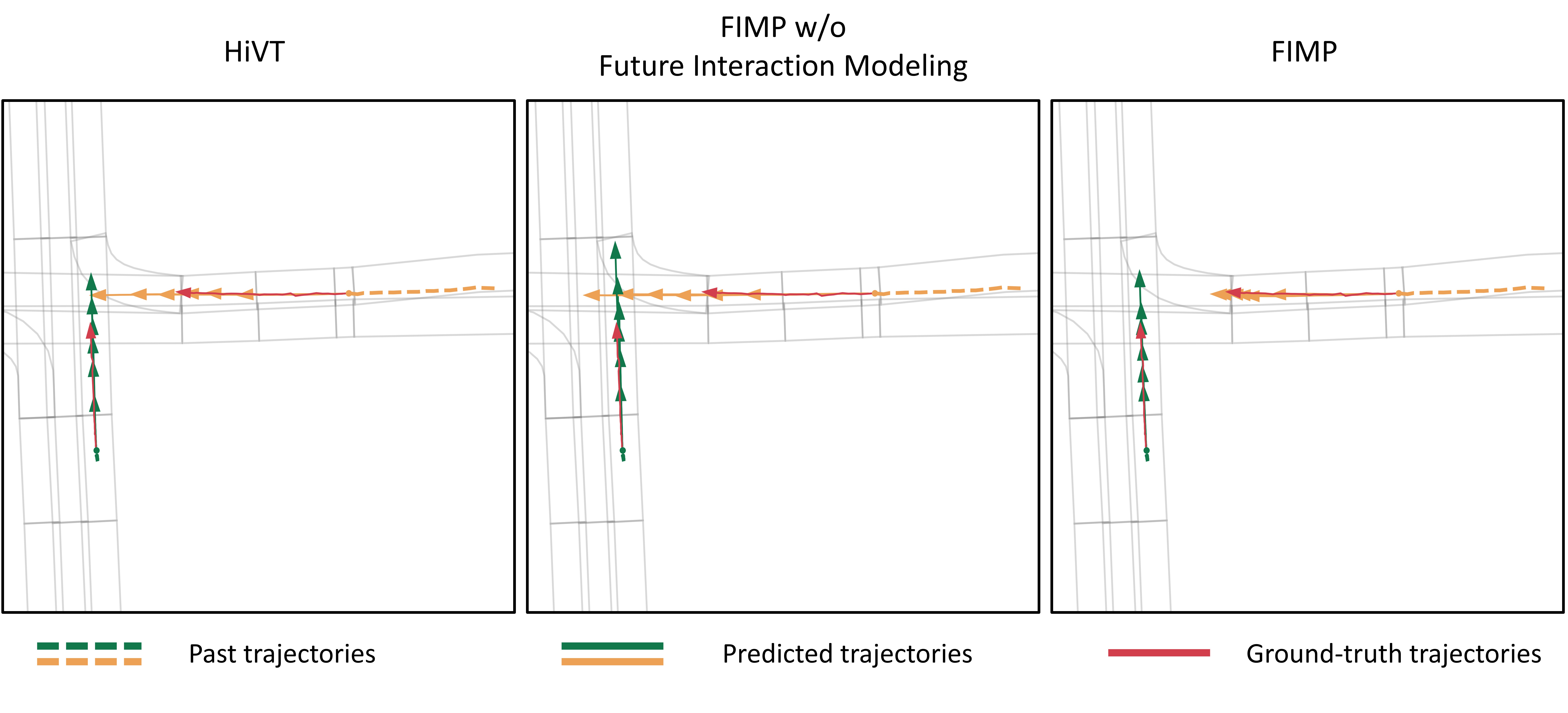}
		
	\end{center}
	\vspace{-0.5cm}
	\caption{In this scenario, the orange agent comes to a halt at an intersection due to the presence of a green agent passing by. As FIMP takes into account the future motions of the interacting agent, it predicts that the orange agent will not approach the green agent, whereas other models predict that the orange agent may attempt to proceed through the green agent's future trajectories.
	}
	
	\label{fig:1}
\end{figure*}
\begin{figure*}[t]
	\begin{center}
		\includegraphics[width=\linewidth]{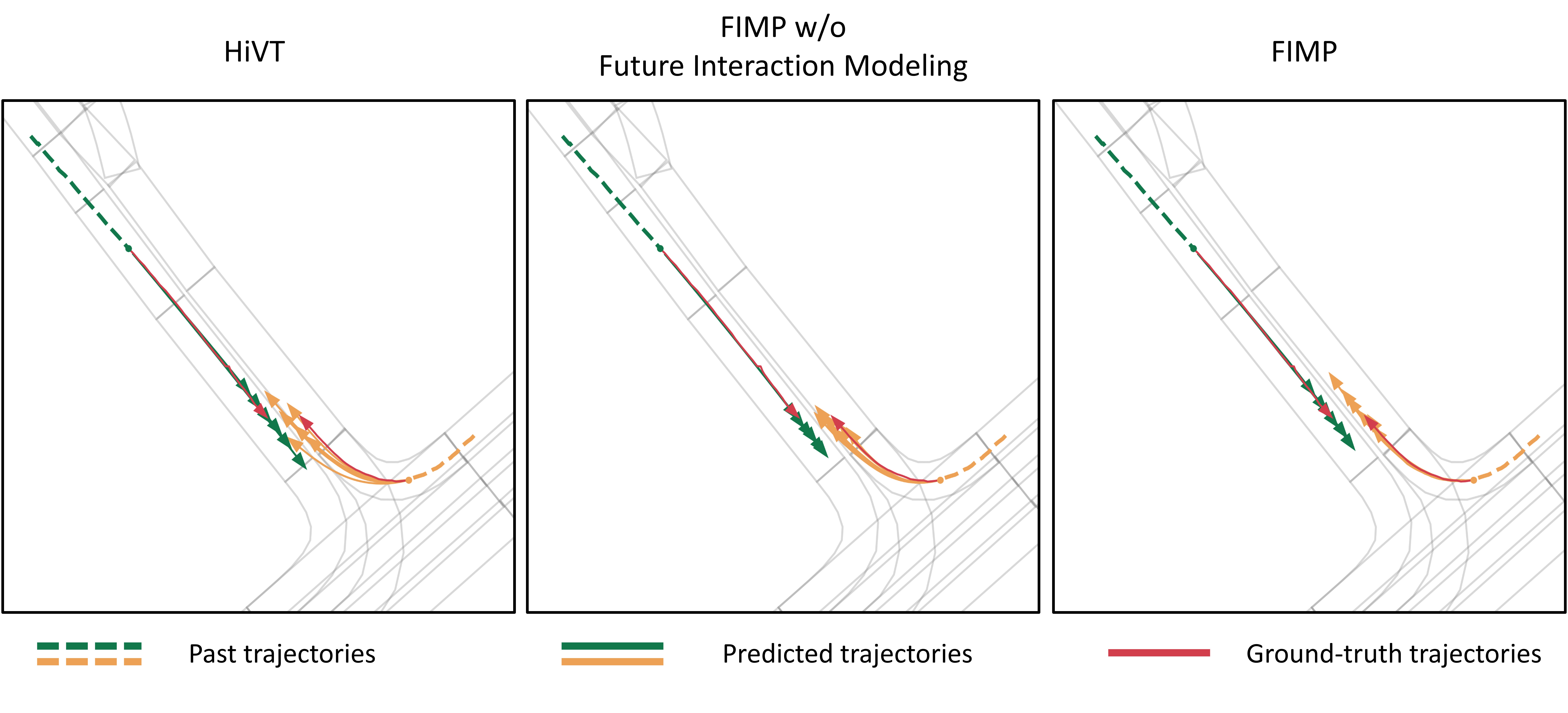}
		
	\end{center}
	\vspace{-0.5cm}
	\caption{In this scenario, two agents are approaching from opposite directions. FIMP's predictions for the orange agent's motion do not cross the centerline, which is a realistic and safe behavior. However, other models predict that the orange agent may cross the centerline, which could result in dangerous situations. HiVT's predictions, in particular, include the possibility of the orange agent colliding with the oncoming green agent.
	}
	
	\label{fig:2}
\end{figure*}
\begin{figure*}[t]
	\begin{center}
		\includegraphics[width=\linewidth]{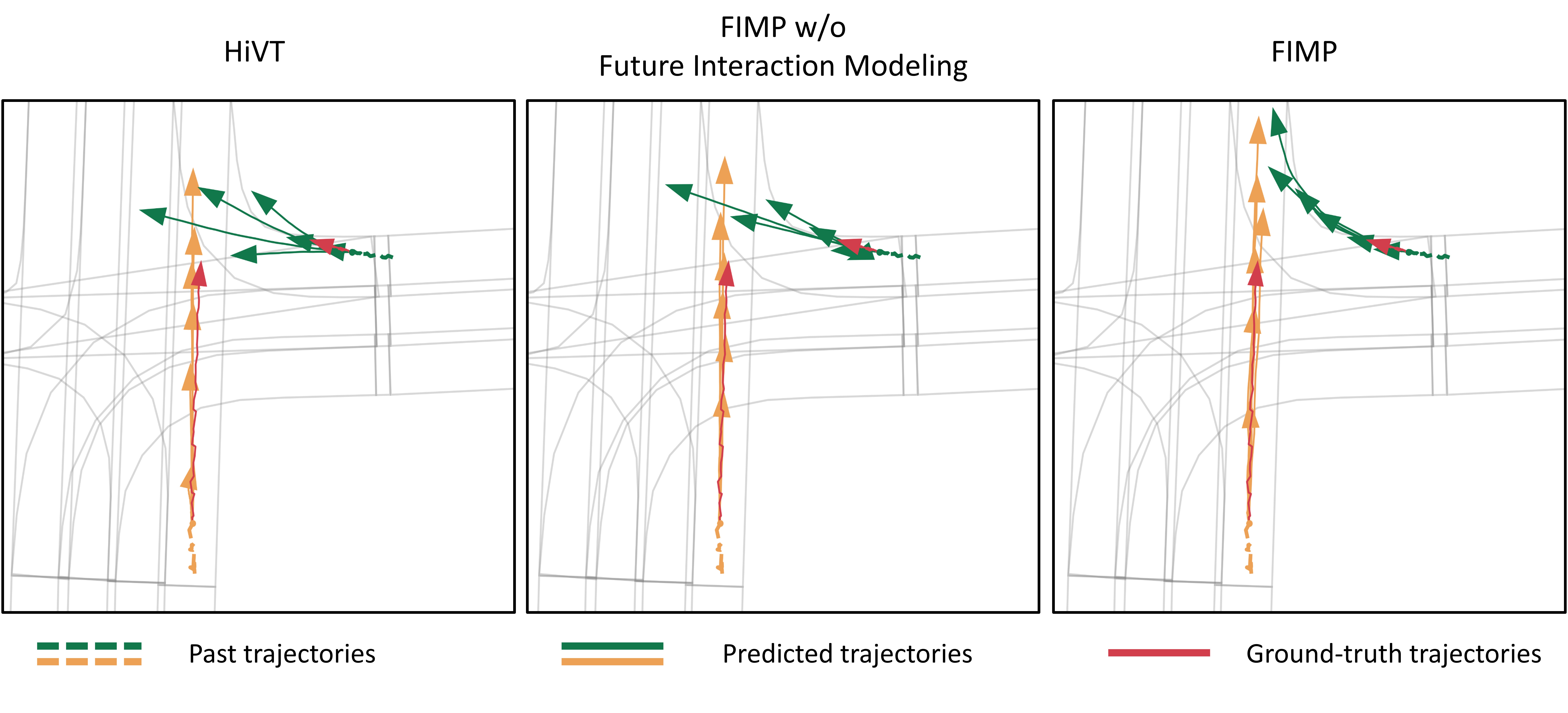}
		
	\end{center}
	\vspace{-0.5cm}
	\caption{In this scenario, one agent is waiting for the other agent to pass by, similar to Figure~\ref{fig:1}. In the modes where the green agent moves without waiting, FIMP's predictions do not result in collisions with the future motions of the orange agent, while the predictions of other models produce overlapping future trajectories.
	}
	
	\label{fig:3}
\end{figure*}
\begin{figure*}[t]
	\begin{center}
		\includegraphics[width=\linewidth]{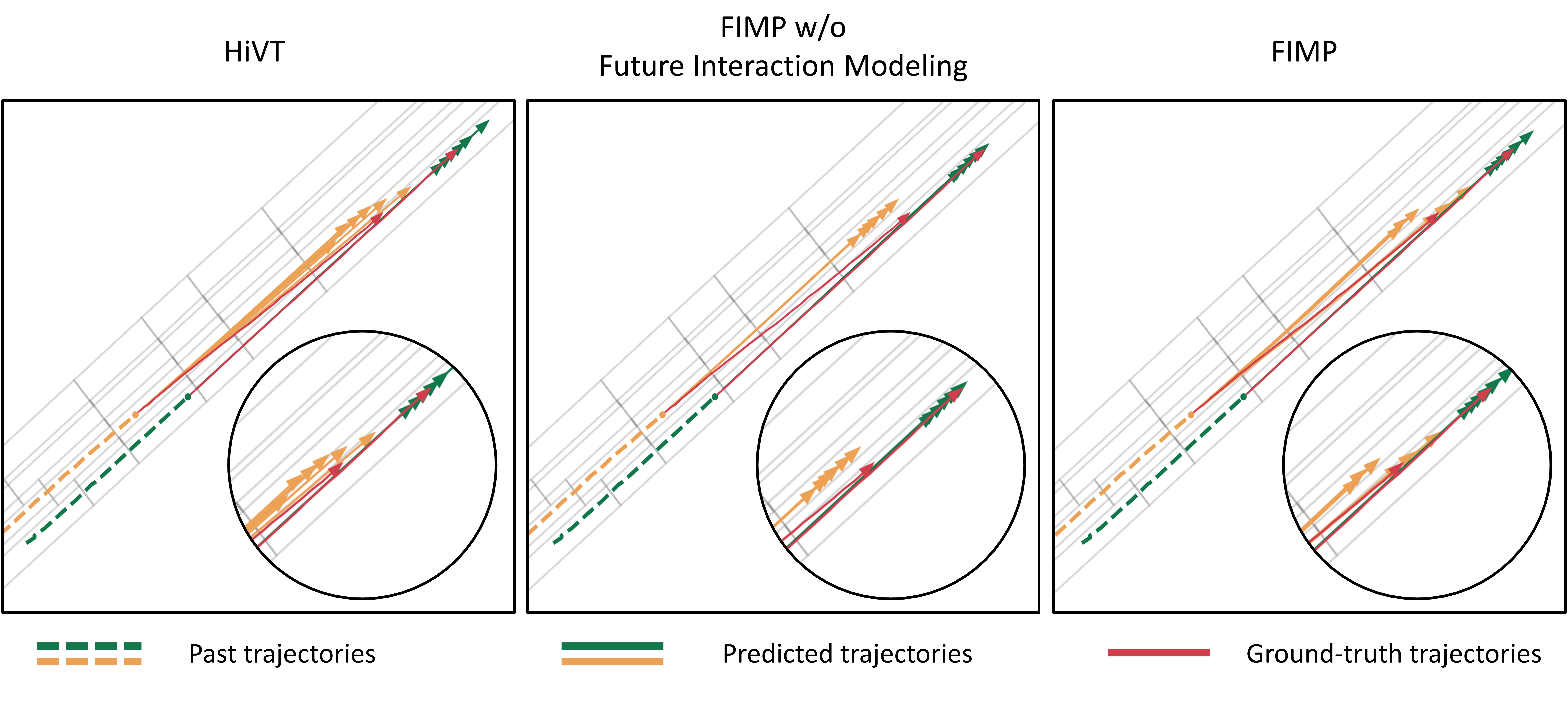}
		
	\end{center}
	\vspace{-0.5cm}
	\caption{This scenario represents a multi-lane road with two agents going to the same direction. As the green agent is ahead at the current frame, the orange agent changes lanes after the green agent passes by. FIMP accurately predicts the timing of the orange agent's lane change by modeling future interaction with the green agent, whereas other models fail to predict the orange agent's future trajectories accurately.
	}
	
	\label{fig:4}
\end{figure*}
\begin{figure*}[t]
	\begin{center}
		\includegraphics[width=\linewidth]{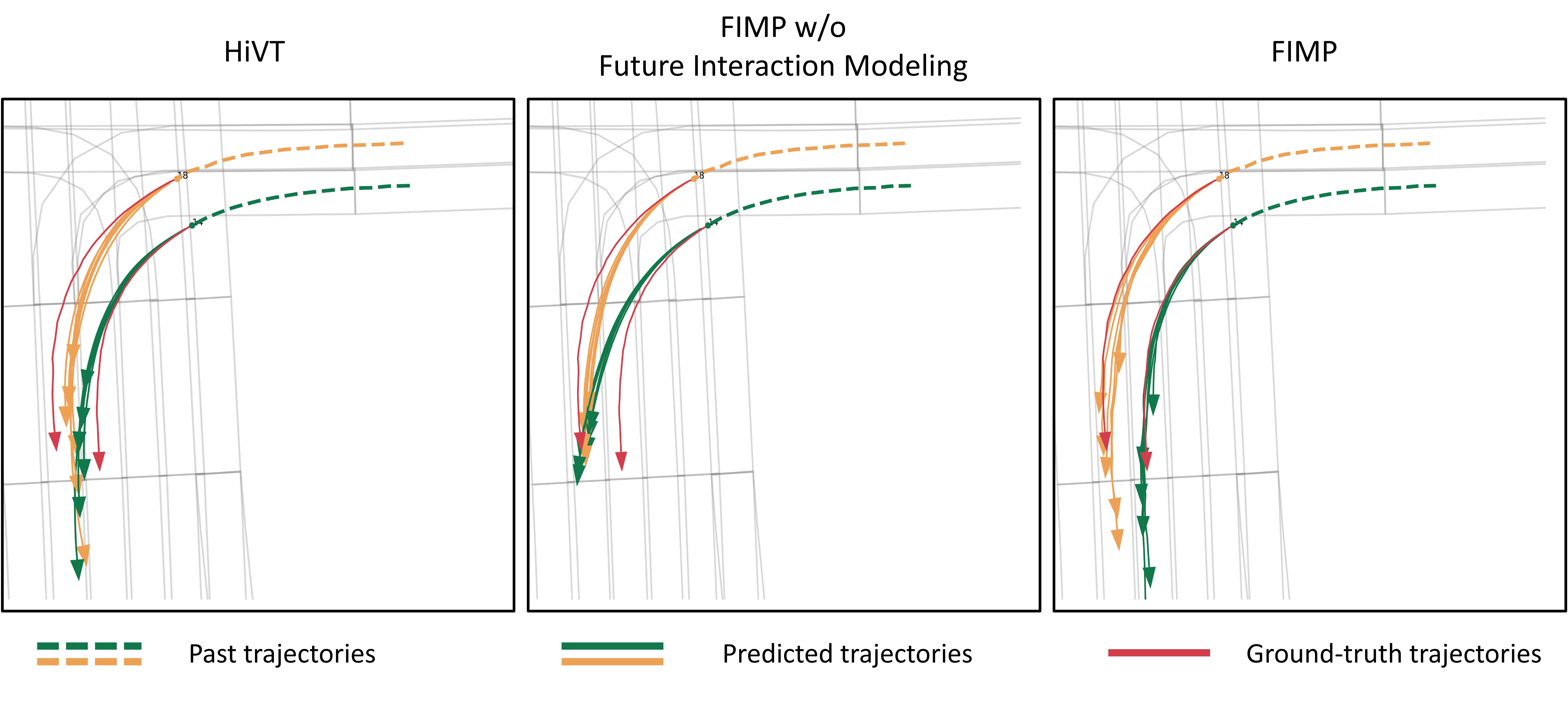}
		
	\end{center}
	\vspace{-0.5cm}
	\caption{This scenario represents the two agents turning left at an intersection. The predictions of FIMP do not collide with the other agent's future trajectories whereas other models predict the overlapping future trajectories which would result in a collision.
	}
	
	\label{fig:5}
\end{figure*}

\addtolength{\textheight}{-0cm}   % This command serves to balance the column lengths

%%%%%%%%%%%%%%%%%%%%%%%%%%%%%%%%%%%%%%%%%%%%%%%%%%%%%%%%%%%%%%%%%%%%%%%%%%%%%%%%

\clearpage
\bibliographystyle{IEEEtran}
\bibliography{egbib}

\end{document}